# Cross-individual Recognition of Emotions by a Dynamic Entropy based Pattern Learning Framework with EEG Features

Xiaolong Zhong and Zhong Yin *

*Abstract*—Using the electroencephalogram (EEG) and machine learning approaches to recognize emotions can facilitate affective human–computer interactions. However, the individual difference of EEG data constitutes an obstacle for cross-individual EEG feature modelling and classification. To address this issue, we propose a deep-learning system denoted as a dynamic entropy-based pattern learning (DEPL) framework to abstract informative indicators pertaining to the neurophysiological features among multiple individuals. DEPL enhanced the capability of representations generated by a deep convolutional neural network by modelling the interdependencies between the cortical locations of dynamical entropy-based features. The effectiveness of the DEPL has been validated with two public databases, commonly referred to as the DEAP and MAHNOB-HCI multimodal tagging databases. Specifically, the leave-one-subject-out training and testing paradigm has been applied. Numerous experiments on EEG emotion recognition demonstrate that the proposed DEPL is superior to classical deep and shallow learning machines, and could learn between-electrode dependencies w.r.t. different emotions, which is meaningful for developing the effective human-computer interaction systems by adapting to human emotions in the real world applications.

This work is sponsored by the National Natural Science Foundation of China under Grant No. 61703277 and the Shanghai Sailing Program under Grant No. 17YF1427000.

X. Zhong is with the School of Optical-Electrical and Computer Engineering, University of Shanghai for Science and Technology, Shanghai, 200093, P. R. China (e-mail: 182560463@usst.edu.cn).

*Corresponding author, Z. Yin is with Engineering Research Center of Optical Instrument and System, Ministry of Education, Shanghai Key Lab of Modern Optical System, University of Shanghai for Science and Technology, Shanghai, 200093, P. R. China (e-mail: yinzhong@usst.edu.cn).

## I. INTRODUCTION

Emotion recognition (ER) plays a significant role in affective human–machine interactions. The purpose of ER is to retrieve the affective status of human beings at a particular point in time given a relevant data recording from an individual. There is a high possibility for applications in medical-care systems, such as active and assisted living module [1], driver-assistance systems [2], early detection of depression [3], and autistic spectral disorders [4]. Feasible clues for an ER task include body gestures [5], facial expressions [6], speech [7], eye blinking [8], and neurophysiological signals [9]–[10]. Among them, neurophysiological signals have a capability to reflect the inner cognitive states of individuals and make impartial ER systems possible [11].

In our study, we concentrated on leveraging the ongoing electroencephalography (EEG) to build a functional ER system. Previous studies on cognitive psychology demonstrated the association between affective information of the human emotional state and the electrical activity of the cerebral cortex. The EEG can be a direct consequence of specific affective stimuli [12]. Nevertheless, extracting emotional indicators from the EEG is difficult because of the individual differences, non-stationarity, and the artifacts induced by eye movement and respiration. A promising solution for the identification of salient information in the EEG associated with human affective responses is to build reliable feature representations and pattern classification models with ML tools.

In particular, there is a statistical significance in the way individuals sense and express their feelings [13]. People may express distinct feelings and cortical responses when exposed to an identical effective stimulus [14]. Therefore, the pattern classifier of interpreting EEG may not yield the right choice with regard to the context of various users of an ER system [15]. Accordingly, many studies tend to focus on the design of specific ER systems. However, this classification model requires a large part of the gathered EEG samples from a specific person, and the recorded EEG could not be utilized for unseen individuals.

Accordingly, studies into cross-individual ER systems based on machine learning models have been investigated. This signifies that a classifier can transfer knowledge pertaining to EEG data distribution among multiple ER users [16]. In such circumstances, the ER model is trained and tested by different individuals. It significantly reduces the time cost for recording EEG signals from the same person given that the size of the available training data is enlarged when multiple users are involved. However, the current cross-individual approach to EEG classification tends to perform poorly compared with the individual-specific approach because of the severe individual differences of EEG data

distributions [17]

To model the mapping between EEG data and emotional states more accurately based on a cross-individual paradigm, we adopted deep-learning (DL) methods [18]. There have been numerous successful utilizations of DL to the broad-scale image, voice, and video data. In contrast to many static images, EEG signals are multichannel, nonlinear time series, and the quantity of instances in numerous public databases is limited, thus making this modality less sufficient for training broad-scale networks that contain millions of parameters. Therefore, current convolutional neural network (CNN) models based on the DL principle have to leverage data augmentation and model regularization techniques. In this study, the CNN is applied as a basis to develop the cross-individual ER system. The CNN-based emotion classifier is trained based on end-to-end learning of the abstract features from the deep-scale original data.

To improve the characterization of the dynamical properties in the EEG features across multiple users of the ER system, we introduce a dynamic entropy-based pattern learning (DEPL) framework to smooth out short-term fluctuations and highlight long-term trends or cycles of feature representations. The DEPL first introduces differential entropy (DE) to characterize EEG signals [19]. The spatial information that is coded in the electrode placement in the cerebral cortex is extracted by transferring DE features to interpretable two-dimensional (2-D) maps. In particular, the squeeze-excitation (SE) block [20] is employed to enhance the capability of deep representations of DE features by modeling the interdependencies between the channels of its informative features. Finally, the CNN classifiers are trained separately based on the classical EEG frequency bands of theta, alpha, beta, and gamma, to predict the final affective states. The public datasets DEAP [21] and MAHNOB-HCI [22] are adopted to validate the effectiveness of the proposed method.

In summary, our main contributions are three folds:

(1) The DEPL deep learning system is first introduced to the EEG based affective computing for cross individual emotion recognition.

(2) A method of salient region extraction based on attention mechanism is designed in the DEPL. The inter-channel dependencies have been evaluated along with the high-level feature representation.

(3) The validation of the DEPL and various deep learning models shows connections between different structures of CNNs and the corresponding generalization capability.

This study is organized as follows. In section II, we introduce related work in the field of EEG-based affective computing. Section III describes the data preprocessing steps of the DEAP and MAHNOB-HCI databases. The details of the proposed DEPL framework are presented in Section IV. The experiments for performance evaluation and comparison of the ER system are given in section V. The discussions of the results and the conclusion of the study are presented in sections VI and VII, respectively.

## II. RELATED WORKS

Most EEG-based-affect classification models label and represent various facets of emotions, and subsequently apply the two-dimensional bipolar affective model based on valence and arousal [23]. Atkinson and Campos [24] used the minimum-redundancy maximum-relevance (mRMR) method for feature selection, and the support vector machine (SVM) for the binary classification of low/high valence and arousal for individual-dependent emotion recognition based on data from the DEAP database. Yoon and Chung [25] presented a classification method based on the Pearson correlation coefficient and Bayesian models that achieved accuracies of 70.9% for valence and 70.1% for arousal. In [26], the level feature fusion (LFF) algorithm was employed to extract emotional EEG features, and the SVM with a Gaussian kernel was used as the classifier. The model obtained an accuracy of 68.8% for valence and 63.6% for arousal for binary classification, and 59.57% for valence and 57.44% for arousal for three-class classification of the MAHNOB database.

In [27], a DL model was used based on the long-short term memory framework to classify low/high valence and arousal based on the EEG raw data from the DEAP database, with accuracies of 85.45% and 85.65%. In [28], a three-dimensional (3-D) CNN-based scheme was applied to classify emotional states, and achieved a mean accuracy of 87.44% for valence and 88.49% for arousal on the DEAP dataset. The training samples were summed by an augmentation process following the superposition of noise on the original EEG sequences.

Most of the existing work on individual, independent ER tasks tends to perform poorly compared with the individual–specific approaches. Li et al. [29] adopted the SVM classifier based on the automatic feature selection methods and the "leave-one-subject-out" verification strategy to evaluate the ER performance on the DEAP database based on which the highest mean recognition accuracy of 59.06% was achieved for binary classification. In [30], a cross-individual ER model was proposed from EEG signals with the use of variational mode decomposition (VMD) as a feature extraction technique and DL networks as the classifier. They obtained accuracies of 61.25% and 62.50% for valence and arousal dimensions on the DEAP database, respectively.

Various EEG-based models have been proposed and applied on ER tasks. Few emotional classification frameworks attempted to exploit the capability of intermediate feature representation of DL methods on cross-individual concept transfers. Moreover, several existing DL models lacked an established dynamical structure to capture a mapping between the current emotional state and the EEG feature in previous time steps. We also notice that the use of entropy measures of recorded EEG signals to distinguish brain emotional states is a powerful tool for feature extraction. The above observations from the literature motivate our DEPL framework in designing the network structure and training algorithms.

## III. DATASETS AND EEG FEATURES

### A. Dataset Description and Data Acquisition

The DEAP is an open-source database recording of multimodal physiological signals with emotional evaluations generated by 32 volunteers with selected video clips. Specifically, this database includes 32-channel EEG and peripheral physiological signals. Each participant was asked to watch 40 trials of music videos (one-minute per video) with different emotional stimuli. The EEG signals of these volunteers were recorded simultaneously. The volunteers then rated the videos on a scale of 1 to 9 in terms of arousal, valence, liking, dominance, and familiarity.

In this study, we only focused on 2-D emotional models where the arousal (ranging from weak to strong) and valence dimensions (ranging from negative to positive) were adopted to generate target-affective states. The 40 stimulus videos included 20 low-arousal/valence clips and 20 high-arousal/valence clips. Only the EEG data in the DEAP were used for deep-network modelling that had been downsampled to 128 Hz. The frequency component (in the range of 4–45 Hz) was preserved by band-pass filtering and ocular artifacts were removed by blind source separation algorithms. The duration of the denoised EEG in each trial was 63 s. Of these, 60 s were experimental data (recorded while watching the video), and 3 s were pre-trial baseline data (before watching), which contained a total number of samples of $(60 + 3) \times 128 = 8064$ for each channel.

TABLE I
DATASET DESCRIPTION

| Feature | DEAP/MAHNOB description |
|---|---|
| number of participants | 32/24 |
| number of videos | 40/20 |
| number of electroencephalographic (EEG) channels | 32 |
| rating scales | valence and arousal |
| rating values | 1–9 |
| sampling rate | 128 Hz |
| duration of experimental signals | 60 s |
| duration of pre-trial baseline signals | 3 s/none |

Similarly, the MAHNOB-HCI database contains EEG, video, audio, gaze, and peripheral physiological recordings from 30 participants. Each participant watched 20 clips extracted from Hollywood movies and video websites. The stimulus videos ranged in duration from 35 s to 117 s. Their perceived arousal and valence levels were also labelled on a discrete scale from 1 to 9. Given that the EEG data were incomplete in the cases of six participants, only the data from 24 participants were available. The sampling frequency of the EEG was also 128 Hz, and the average potential was used as a reference value to measure the EEG potential difference. High-pass filters were employed to perform artifact removal. Specifically, for each trial of the EEG, a 60 s segment from the first 5 s to the 65 s time-point were used in the following analysis. We then divided these EEG recordings to high-class (ratings 6–9) and a low-class (ratings 1–5) recordings. Table I shows a summary of the DEAP and MAHNOB-HCI datasets.

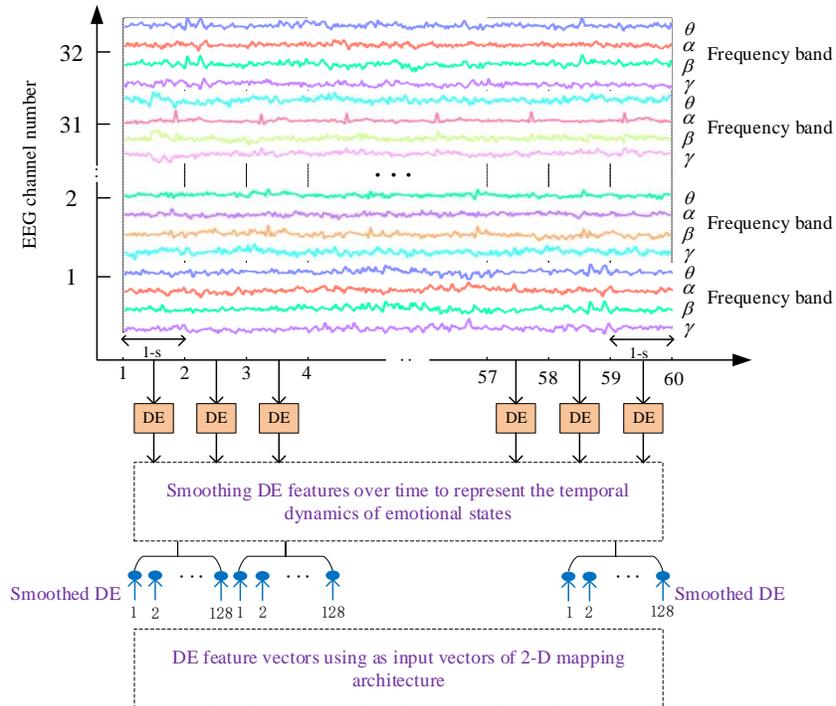

Fig. 1. An illustration of mapping electroencephalographic (EEG) signals over time to differential entropy (DE) feature vectors.

## B. Computation of Differential Entropy

Applying complexity measures with entropy-based pattern learning has been certified as one of the leading technical means for EEG-based emotion recognition [31]. Entropy measures can be exploited to quantify the nonlinearity, uncertainty, and non-stationarity of neurophysiological signals [32]–[33]. Additionally, there is plenty of persuasive evidence that illustrates that the entropy measures have a strong ability to extract regularity information from EEG 8signals indicating clinical significances [34]. In this study, the DE is presented to characterize the affective clues in an EEG recording, and is defined as follows,

$$h(X) = -\int_S f(x)\log(f(x))dx. \quad (1)$$

In (1), $S$ is the support set of the random variable, $X$ is a continuous random variable of the observed value $x$, and $f(X)$ is the probability density function of $X$. The value $x$ is drawn from a normal distribution with a zero mean, i.e. $X \sim \phi(x) = 1/\sqrt{2\pi\sigma^2}\exp(-x^2/2\sigma^2)$. Although the raw EEG signals did not fit any standard distributions, it can be observed by the Kolmogorov–Smirnov (K–S) statistic that the decomposed EEG signals are Gaussian distributed when they are split into the sub-bands after band-pass filtering. Therefore, calculating the DE value can be expressed as,

$$h(X) = -\int \frac{1}{\sqrt{2\pi\sigma^2}} e^{-\frac{(x-\mu)^2}{2\sigma^2}} \log(\frac{1}{\sqrt{2\pi\sigma^2}} e^{-\frac{(x-\mu)^2}{2\sigma^2}})dx. \quad (2)$$

Before applying (2), each EEG segment is decomposed with Butterworth filters in four classical bands, i.e., theta, alpha, beta, and gamma, as shown in Fig. 1. The theta (4–7 Hz) component is active in light-sleep patterns [35], while the alpha (8–13 Hz) is activated in a relaxed state by closing the eyes [36]. The beta (14–30 Hz) band constitutes the "active thinking and reasoning wave" band [37], and the Gamma band (31–45 Hz) is related to bursts of perceptive and advanced information processing [38]. Consequently, in a fixed frequency band $i$, the differential entropy $h_i$ is defined based on (2) as follows,

$$h_i = \frac{1}{2}\log(2\pi e \sigma_i^2). \quad (3)$$

where $\sigma_i^2$ denotes the signal variance.

*C. Data Processing and Feature Extraction*

The data pre-processing and feature extraction processes can be summarized according to the procedure EEG_PRE in Table II. According to Yang et al. [39], the DE indicators of the EEG in baseline conditions can improve the inter-emotion discriminant capability of the feature space. Hence, the differences between the EEG at various experimental conditions and the baseline recording are computed. For each participant, all the 32 channels were decomposed in the various bands $(\theta, \alpha, \beta, \gamma)$. Subsequently, the baseline EEG with the length of 3 s is extracted with a 128 point, non-overlapped window from the component in each frequency band. Thus, the baseline EEG data of each participant is converted to a 40 × 32 × 4 × 384 (trials × channels × bands × sample) tensor. The 384 point data of each channel is then divided into three epochs, and the DE feature is calculated over each 128 point epoch. Four DE feature vectors can be computed and denoted by $\mathbf{X}_{DE}^{base(\theta, \alpha, \beta, \gamma)} \in \mathbb{R}^{32}$. The next step involved the concatenation of all four vectors according to the order of the theta, alpha, beta, and gamma bands into a vector $\mathbf{X}_{DE}^{base} \in \mathbb{R}^{128}$. Finally, the mean of the three data-windows was computed to represent the DE feature of pre-trial baseline signals, i.e.

$$\mathbf{X}_{DE} = \frac{1}{3}\sum_{j=1}^{3}\mathbf{X}_{DE}^{base}(j). \tag{4}$$

The experimental signals (duration of 60 s) were pre-processed with the same operation. The EEG data of each subject were converted to a 40 × 32 × 4 × 7680 (videos × channels × bands × sample) tensor. According to Wang et al. [40], a window with a size of 1 s can be suitable so that the 7680 point data of each channel was divided into 60 epochs. The total number of EEG epochs of each participant was equal to 40 × 60 = 2400 with the dimensionality of 128 (samples) × 32 (channels) × 4 (bands). Subsequently, the corresponding DE feature vector of each epoch $\mathbf{X}_{DE}^{exper(\theta, \alpha, \beta, \gamma)} \in \mathbb{R}^{32}$ was calculated, and the features of the four bands were concatenated as a new vector $\mathbf{X}_{DE}^{exper} \in \mathbb{R}^{128}$.

TABLE II
PROCEDURE FOR DYNAMICAL FEATURE SMOOTHING

| | $\mathbf{X}_{DE}$ = **EEG_PRE**( $\mathbf{X}$ ) |
|---|---|
| **input:** | an epoch of EEG signals $\mathbf{X}$ |
| **output:** | the DE features of an epoch $\mathbf{X}_{DE}$ |
| 1 | decompose $\mathbf{X}$ into $\mathbf{X}^{base}$ and $\mathbf{X}^{exper}$ |
| 2 | **for** $i$ = 1 to 32 |
| 3 | decompose $\mathbf{X}_i^{base}$ into $\mathbf{X}_i^{base(\theta,\alpha,\beta,\gamma)}$ |
| 4 | decompose $\mathbf{X}_i^{exper}$ into $\mathbf{X}_i^{base(\theta,\alpha,\beta,\gamma)}$ |
| 5 | **for** $k$ = 1 to 4 |
| 6 | decompose $\mathbf{X}_{i,k}^{base(\theta,\alpha,\beta,\gamma)}$ as $\mathbf{X}_{i,k}^{base(\theta,\alpha,\beta,\gamma)}(j)$. |
| 7 | compute $\mathbf{X}_{DE,i,k}^{base}$ with Eqn. (4) |
| 8 | decompose $\mathbf{X}_{i,k}^{exper(\theta,\alpha,\beta,\gamma)}$ as $\mathbf{X}_{i,k}^{exper(\theta,\alpha,\beta,\gamma)}(j)$ |
| 9 | compute $\bar{\mathbf{X}}_{DE,i,k}^{exper}$ with Eqn. (5) |
| 10 | $\mathbf{X}_{DE,i,k} = \bar{\mathbf{X}}_{DE,i,k}^{exper} - \mathbf{X}_{DE,i,k}^{base}$ |
| 11 | merge $\mathbf{X}_{DE,i,k}$ into $\mathbf{X}_{DE}$ |
| 12 | **return** $\mathbf{X}_{DE}$ |

IV. DYNAMICAL ENTROPY-BASED PATTERN LEARNING

The ER framework designed based on the DEPL method is shown in Fig. 2. The leave-one-subject-out paradigm has been used to validate the binary classification performance. The DEPL consists of four modules, i.e. dynamical feature smoothing, 2-D feature mapping, the deep network for feature abstraction, and channel feature re-calibration.

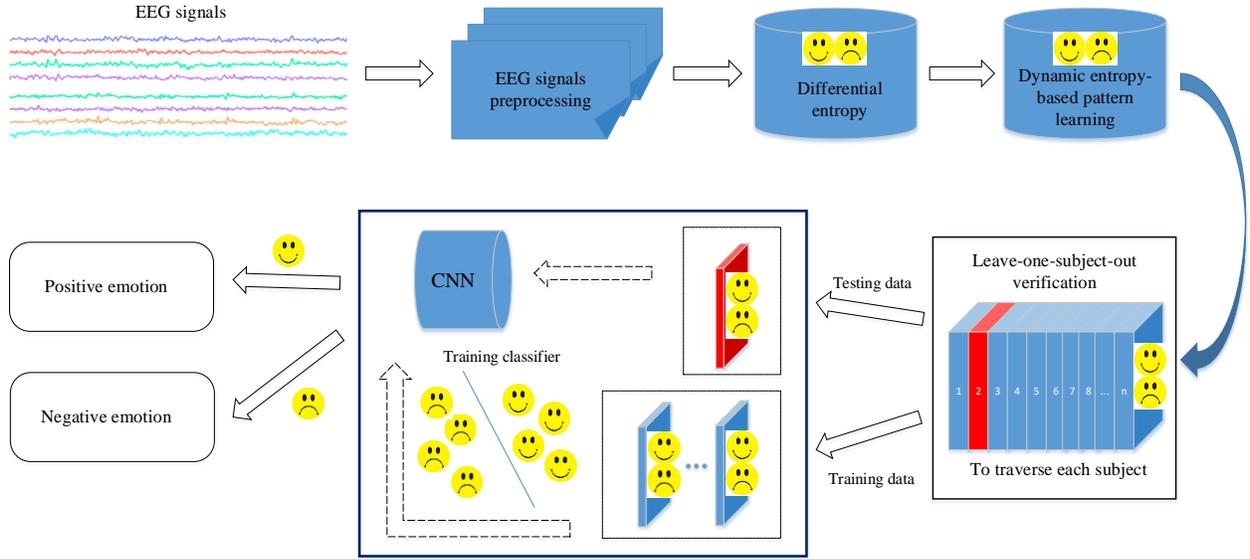

Fig. 2. Dynamic entropy-based pattern learning to enable cross-individual emotion recognition based on EEG signals.

## A. Dynamical Feature Smoothing

The analysis of the experimental EEG segments that have been observed at different time steps led to a dynamical learning problem. The correlation introduced by the sampling of adjacent feature vectors with respect to time restricted the applicability of many conventional statistical methods that depended on the assumption that the observations were independent and identically distributed. Owing to the non-stationarity of the EEG signals, the emotional states estimated by the DE features at the current time step were associated with past affective experiences. Consequently, it was necessary to map DE features over time to track the dynamical characteristics of the EEG distribution.

To this end, we used a dynamic entropy indicator to smooth the DE features, obtained their temporal profiles, and built a novel framework of dynamical entropy-based pattern learning (DEPL) to enable individual-independent ER tasks with EEG features. Given that the DE sequences have difficulty in reflecting the trend of events because of the periodic variation and the effect of stochastic volatility, the DEPL employed the average of historical DE sample values as a feature prediction. Alternatively, the functionality of the DEPL can be regarded as a low-pass filter applied to the time course of the feature. Thus, the DE indicator at the current time step is related to that at all previous time steps, i.e. each second of data per video,

$$\bar{\mathbf{X}}_{DE}^{\text{exper}}(n) = \frac{1}{d}[\mathbf{X}_{DE}^{\text{exper}}(n) + \mathbf{X}_{DE}^{\text{exper}}(n-1) + ... + \mathbf{X}_{DE}^{\text{exper}}(n-d+1)]. \tag{5}$$

In (5), $\mathbf{X}_{DE}^{\text{exper}}(i)$ denotes the DE feature value at time step $i$ of each trial with delay $d$. The DE difference between the experimental and baseline EEG is calculated to represent the emotional indicator of that epoch,

$$\mathbf{X}_{DE} = \bar{\mathbf{X}}_{DE}^{\text{exper}} - \mathbf{X}_{DE}^{\text{base}}. \tag{6}$$

Owing to the lack of a baseline signal in the MAHNOB dataset, (6) is only applied on the DEAP dataset. Table II summarizes the procedure of the dynamical feature smoothing.

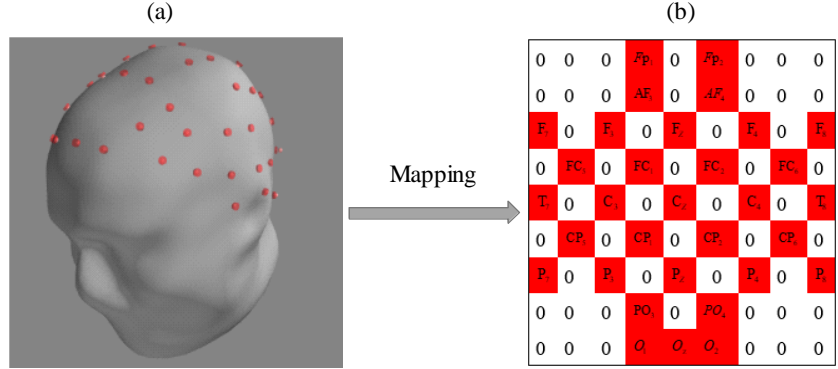

Fig. 3. Construction of the two-dimensional (2-D) plane.

## B. Two-dimensional Feature Mapping

Aggregating the extracted DE indicators for all channels to generate a feature vector is conventionally used in EEG feature analysis. Nevertheless, it does not consider the spatial structure of the cortical areas. To this end, we transform the DE indicators into a 2-D image to keep the consistency of local spatial information among adjacent electrodes. Specifically, the one-dimensional (1-D) feature vector $\bar{\mathbf{X}}_{DE}^{\text{exper}}$ is transformed as a 2-D matrix with the size of $h \times l$, where $h = l = 9$ are the numbers of the vertical and horizontal coordinates of the electrodes. The corresponding feature matrix of the $\bar{\mathbf{X}}_{DE}^{\text{exper}}$ within the frequency bands $i \in \{\theta, \alpha, \beta, \gamma\}$ is denoted as $\mathbf{F}^i \in \mathbb{R}^{h \times l}$. Zeroes are used to fill the DEs from channels that are unused in the experiments. Finally, we are able to acquire four 2-D matrices for each epoch, as shown in Fig. 3.

## C. Deep Network for Feature Abstraction

To preserve the local spatial structure of the EEG features, the deep CNN can be applied on the 2-D-like frame in a functionality of image recognition. In this study, our CNN model is developed based on the LeNet-5 [41], which was the primary models for all types of CNN applications in the domain of computer vision, as shown in Fig. 4(a).

Given an input EEG feature map indicated by $\mathbf{x}$, an abstracted feature at the $i^{th}$ row and $j^{th}$ column in a convolutional layer indicated by $y_{ij}$ can be calculated by applying the spatial convolution operation,

$$\mathbf{y}_{ij}(k) = f(\mathbf{w}(k)^T \mathbf{x}_{ij}(k) + b_{ij}). \tag{7}$$

In (7), $\mathbf{w}(k)$ is the connecting weight of a convolution kernel with a size of $k \times k$, $\mathbf{x}_{ij}(k)$ indicates the input with a $k \times k$ receptive field centered at the $i^{th}$ row and the $j^{th}$ column, and $b_{ij}$ is the bias. The term $f(\cdot)$ denotes the activation function. Normally, $k$ is a predefined integer, and is fixed throughout the training/testing process.

In all the types of activation functions, the rectified linear unit (ReLU) [42] was used, and was the most successful. Conversely, according to the conclusion inferred by Ramachandran et al. [43], the Swish activation can help improve the convergence speed of the training. Thence, we used the Swish activation function instead of the ReLU. The Swish function is a smooth, non-monotonic function defined as $f(x) = x \cdot (1/(1+e^{-x}))$.

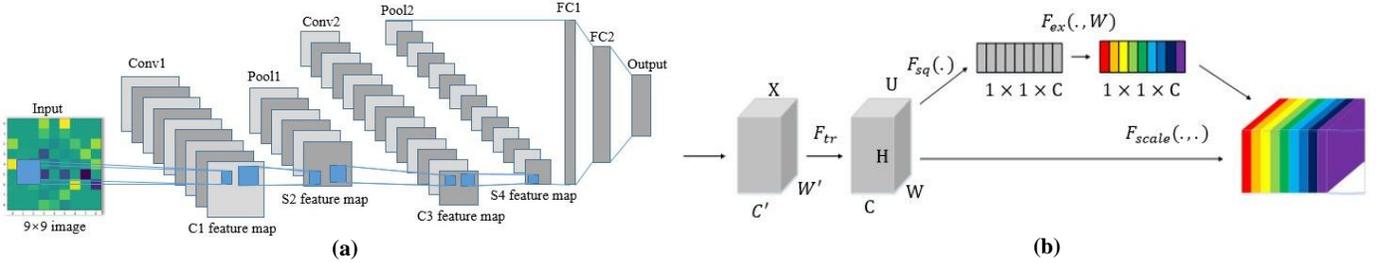

Fig. 4. Basic architecture of (a) the LeNet-5 and (b) the squeeze-excitation building block.

*D. Channel Feature Recalibration*

To strengthen the abstraction capability of a CNN, prior research had indicated the profits of heightening the spatial encoding [44]. Recent studies have revealed that the deep feature representations can be boosted by introducing learning mechanisms into the network that support catch-spatial correlations between features. Therefore, we employ the squeeze-excitation (SE) block to allow the network to perform channel feature recalibration (see Fig. 4(b)).

At any given transformation $\mathbf{F}_{tr} : \mathbf{X} \to \mathbf{U}$ with $\mathbf{X} \in \mathbb{R}^{W' \times H' \times C'}$ and $\mathbf{U} \in \mathbb{R}^{W \times H \times C}$, e.g. a convolution, we would be able to produce a relevant SE block to perform the feature re-calibration. First, the features $\mathbf{U}$ are performed as a squeeze action that generate a channel descriptor based on the fusion of the feature maps spanning in $H \times W$. This descriptor intends to generate an embedding of the global spatial information of channel-wise feature responses so that the functional signal from the global receptive field can be used by its lower layer. The information aggregated in the squeeze operation is followed by an excitation action. It is in the form of a simple self-gating mechanism that embeds as an input, and generates a set of modulation weights for each channel. Applying these weights to the feature map $\mathbf{U}$ produces the output of the SE block that can be fed immediately to the subsequent network layers.

In a SE block, the convolutional operator $conv(\cdot,\cdot)$ can be defined as,

$$\mathbf{u}_c = conv(\mathbf{v}_c, \mathbf{X}) = \sum_{s=1}^{C'} \mathbf{v}_c^s * \mathbf{x}^s. \tag{8}$$

In the equation, $*$ represents the operation of convolution, and $\mathbf{v}_c = [\mathbf{v}_c^1, \mathbf{v}_c^2, ..., \mathbf{v}_c^{C'}]$ denotes the parameters of the *c-th* filter with the input $\mathbf{X} = [\mathbf{x}^1, \mathbf{x}^2, ..., \mathbf{x}^{C'}]$ and the output $\mathbf{u}_c \in \mathbb{R}^{H \times W}$. The term $\mathbf{v}_c^s$ denotes a 2-D spatial kernel indicating a single channel of $\mathbf{v}_c$ that works on the relevant channel of $\mathbf{X}$. The sensor $\mathbf{V} = [\mathbf{v}_1, \mathbf{v}_2, ..., \mathbf{v}_C]$ represents the learnt group of the filter kernel and $\mathbf{U} = [\mathbf{u}_1, \mathbf{u}_2, ..., \mathbf{u}_C]$ denotes the outputs of the mapping $\mathbf{F}_{tr}$. Next, the global spatial information is squeezed into a channel descriptor $\mathbf{z} \in \mathbb{R}^C$ with the c-*th* element denoted by $\mathbf{z}_c$,

$$\mathbf{z}_c = \mathbf{F}_{sq}(\mathbf{u}_c) = \frac{1}{H \times W} \sum_{i=1}^{H} \sum_{j=1}^{W} u_c(i, j). \tag{9}$$

The excitation operation is then performed with a sigmoid activation and the final output of the block is obtained by rescaling $\mathbf{U}$ with the activations $\mathbf{s}$,

$$\mathbf{s} = \mathbf{F}_{ex}(\mathbf{z}, \mathbf{W}) = f\left(\mathbf{W}_2 g\left(\mathbf{W}_1 \mathbf{z}\right)\right). \tag{10}$$

$$\tilde{\mathbf{x}}_c = \mathbf{F}_{scale}(\mathbf{u}_c, s_c) = s_c \cdot \mathbf{u}_c. \tag{11}$$

In the above equations, $g$ and $f$ denote the ReLU and sigmoid functions with weights of $\mathbf{W}_1 \in \mathbb{R}^{C/r \times C}$ and $\mathbf{W}_2 \in \mathbb{R}^{C \times C/r}$, respectively. The term $\mathbf{F}_{scale}(\mathbf{u}_c, s_c)$ denotes channel-wise multiplication between the scalar $s_c$ and the feature map $\mathbf{u}_c \in \mathbb{R}^{H \times W}$.

*E. Training and Implementation*

Given that the final goal of the ER task is to identify a decision function $f_s(\mathbf{S}): \mathbf{S} \to \mathbf{y}$, we trained our network that was parameterized by $\boldsymbol{\theta}$, and minimized a training loss function $\mathcal{L}$:

$$\boldsymbol{\theta}^* = \arg\min_{\boldsymbol{\theta}} \mathcal{L}\left(\mathbf{y}, \hat{\mathbf{y}} = f_s(\mathbf{S}, \boldsymbol{\theta})\right), \tag{12}$$

where $\mathcal{L}$ refers to a cross-entropy cost calculated by,

$$\mathcal{L} = -\frac{1}{n}\sum_{i=1}^{n}\sum_{j=1}^{q} y_j^{(i)} \log \hat{y}_j^{(i)}. \tag{13}$$

In this way, our training target can be set to make the predicted probability distribution $\hat{y}^{(i)} = g_{sm}(o^{(i)})$ as close to the ground truth label $y^i$ as possible. The operator $g_{sm}(\cdot)$ denotes the SoftMax operation,

$$g_{sm}(o_i) = \frac{\exp(o_i)}{\sum_{k=1}^{K}\exp(o_k)}. \tag{14}$$

TABLE III
PROCEDURE FOR FEATURE ABSTRACTING AND RECALIBRATION

| | |
|---|---|
| **input:** | labelled instances $\{(\mathbf{x}^{(n)}, \mathbf{y}^{(n)})\}$ |
| | initialized weight $\mathbf{w} = \mathbf{w}_0$ and bias |
| **output:** | learnt model parameters $\mathbf{w}, \mathbf{b}$ |
| 1 | iteration←0 |
| 2 | **while** *number of epochs* > iteration **do** |
| 3 | iteration←iteration + 1 |
| 4 | **for** *each batch* **do** |
| 5 | convolutional neural network (CNN) forward pass → $\hat{y}^{(i)}$ |
| 6 | compute cross-entropy loss $\mathcal{L}$ via (13) |
| 7 | compute objective derivative as follows |
| 8 | **for** *each layer l* **do** |
| 9 | **if** *this layer is a fully connected layer* **do** |
| 10 | $\frac{\partial \mathcal{L}}{\partial \mathbf{w}^l} \leftarrow \boldsymbol{\delta}^l (\mathbf{a}^{l-1})^T, \frac{\partial \mathcal{L}}{\partial \mathbf{b}^l} \leftarrow \boldsymbol{\delta}^l$ |
| 11 | **else if** *this layer is convolutional layer* **do** |
| 12 | $\frac{\partial \mathcal{L}}{\partial \mathbf{w}^l} \leftarrow \boldsymbol{\delta}^l * f(\mathbf{z}^{l-1}), \frac{\partial \mathcal{L}}{\partial \mathbf{b}^l} \leftarrow \sum_x \sum_y \boldsymbol{\delta}^l$ |
| 13 | update CNN weights as follows |
| 14 | $\mathbf{w}^l = \mathbf{w}^l - \frac{\eta}{b}\sum \frac{\partial \mathcal{L}}{\partial \mathbf{w}^l}$ |
| 15 | $\mathbf{b}^l = \mathbf{b}^l - \frac{\eta}{b}\sum \frac{\partial \mathcal{L}}{\partial \mathbf{b}^l}$ |
| 16 | **return** all $\mathbf{w}^l$ and $\mathbf{b}^l$ as $\mathbf{w}, \mathbf{b}$ |

Note: $\boldsymbol{\delta}^l$ represents the error produced by layer $l$. The term $\mathbf{z}^{l-1}$ represents the input of neurons of layer $l$-1. $\mathbf{a}^{l-1}$ represents the output of neurons of layer $l$-1, and $f$ represents the activation function of the neurons of the convolutional layer.

In the DEPL framework, the first convolutional layer filters the 9 ×9, 2-D input array that corresponded to each frequency band based on the use of 100 kernels with three rows and three columns. The next layer is a max-pooling layer with a pooling size of 2 ×2 and a stride of 2 pixels. Another convolutional layer was then added with 100 filters, and a 3×3 kernel was used as the input. Followed by this layer was another max-pooling layer with the same hyper-parameters. At the end of the network, three fully connected layers with 120, 120, and 2 neurons are added for supervised emotion recognition on the binary arousal and valence levels.

Owing to the fact that the limited number of samples may lead to overfitting, we introduce an L2 regularization term into the error signal of every convolution layer and employed the dropout mechanism at all the fully connected layers. In addition, the batch normalization (BN) was implemented with the SE block between the convolutional layer and the max-pooling layer. The model training process is detailed in Table III.

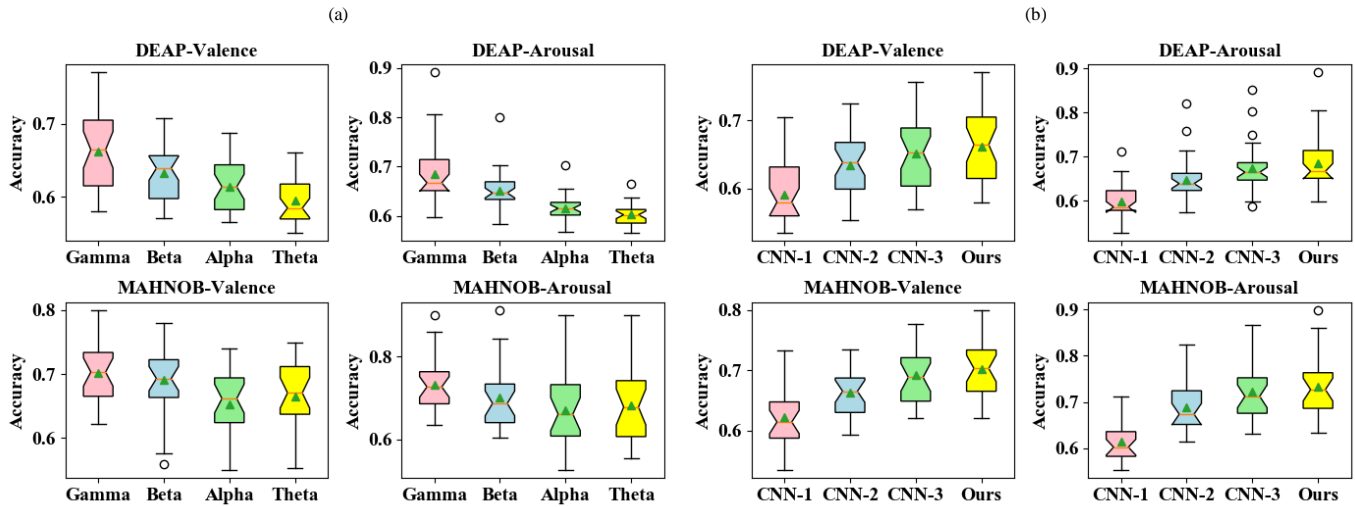

Fig. 5. Box plots of the average participant classification accuracies of two databases. Subfigure (a) represents the classification accuracies of theta, alpha, beta, gamma frequency features for the DEPL model. Subfigure (b) compares the classification accuracy of the dynamic entropy-based pattern learning (DEPL) with gamma features and other deep learning models.

## V. EXPERIMENTS AND RESULTS

### A. Experimental Setup

The leave-one-subject-out paradigm with z-scored features has been used in the following sections to facilitate cross-individual EEG classification in which a single participant obtained from the entire EEG database was used as the test participant, while the remaining data were used in the training process. This cross-validation process was repeated until each participant was used as a test participant.

We implemented the DEPL with the Keras framework libraries in Python and trained it on a Tesla T4 graphics processing unit (GPU) based on Google's cloud platform. The truncated normal distribution function was used to initialize the weight of the kernels, and the Adam optimizer was adopted to minimize the cross-entropy loss function. The initial learning rate was 1.0e-05. The key probability of the dropout operation was 0.6. The penalty strength of the L2 regularization was 0.6. We used 100 epochs and trained our model with batches, which contained 32 experiments each. Finally, we fine-tuned the network to obtain the final classification model. The classic method used to compare with our method was implemented with scikit-learn, and was trained with a laptop computer that had a Windows 10® operation system, an Intel®i5 central processing unit (CPU) at 1.60 GHz, and 4 G configurations.

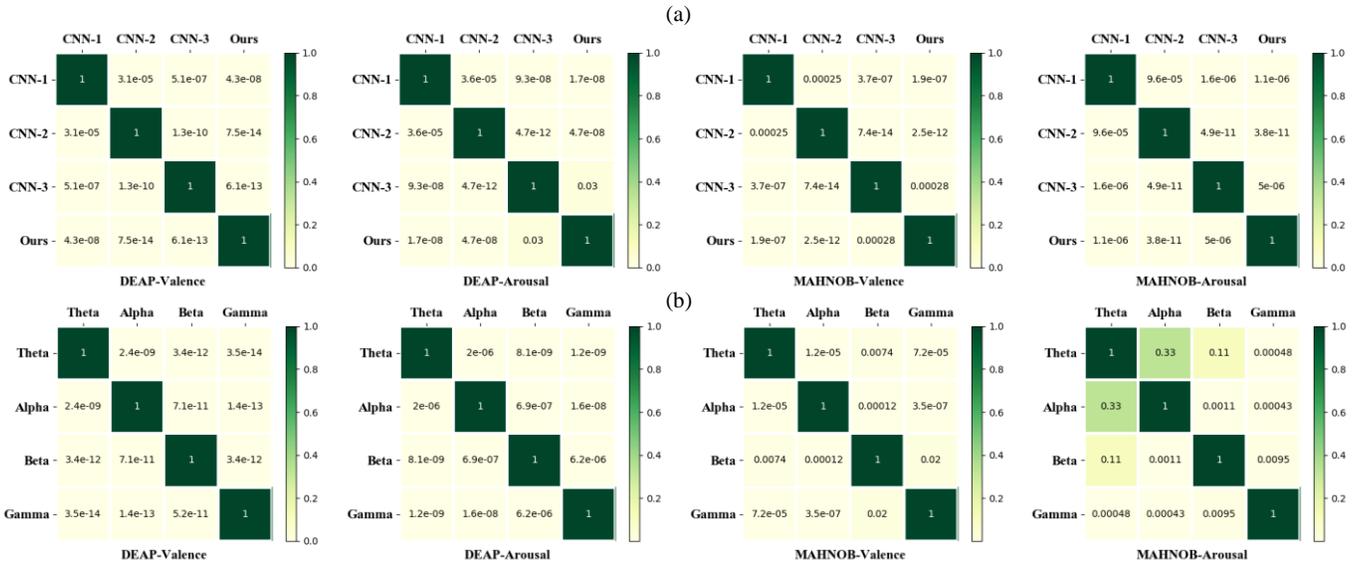

Fig. 6. Heat-map plots of the *p*-values of paired *t*-test between (a) the four EEG bands of the different DEPL rhythms and (b) the four different deep models.

## B. Comparison of Different CNN Structures

Before implementing the DEPL, the optimal frequency band had been examined in Fig. 5(a). For the DEAP database, the highest and the lowest accuracies were achieved by the gamma and theta band features, respectively. For the MAHNOB database, the highest and the lowest accuracies were achieved by the gamma and alpha band features, respectively. The arousal and valence dimensions of the same database possessed similar performance patterns.

In Fig. 6(b), we utilized paired *t*-tests to explore whether the improvements between various bands were significant or not. It is shown that the average accuracies of participants between each two bands were significantly different for the DEAP database. The observation indicates that using gamma features significantly improved the DEPL performance compared with theta, alpha, or beta features ($p < 0.001$). Similar observations were found in the case of the MAHNOB database with $p < 0.05$. Therefore, we implemented the DEPL model with the use of gamma features.

We also compared the current network structure of the DEPL with three popular deep CNN architectures, i.e., the CNN models with ReLU function without feature smoothing (denoted as CNN-1), the CNN model with ReLU function, feature smoothing (denoted as CNN-2), and the CNN model with the ReLU function, SE blocks, and feature smoothing (denoted as CNN-3). The comparison of the accuracy and the *p* values showed that the current network structure possessed the highest participant average performance.

## C. Comparison with Shallow Learning Machines

Several state-of-the-art machine-learning classifiers were considered as baseline classifiers. These included SVM, logistic regression (LR), extreme gradient boosting (XGBoost), gradient boosting decision tree (GBDT), K-nearest neighbors (KNN), decision tree (DT), random forest (RF) and naive Bayes (NB) models. All hyper-parameters of the shallow learning machines have been optimized in detail, and the optimal value is shown in Table IV. Note that the DEPL model applied the optimal structure as shown in the previous section.

TABLE IV
HYPER-PARAMETER SETTINGS FOR SHALLOW LEARNING MACHINES

| Classifier | Hyper-parameter settings |
|---|---|
| KNN | $k = 20$ |
| NB | Gaussian distribution |
| RF | number of the estimators = 200 |
| AdaBoost | number of the estimators = 200, maximum depth = 24 |
| XGBoost | number of the estimators = 200, maximum depth = 22 |
| DT | maximum depth = 7, minimum samples in the leaf node = 12 |
| SVM | regularization parameter = 6 with Gaussian basis kernel |
| GBDT | number of the estimators = 200, maximum depth = 16 |
| LR | generalized linear models |

Table V lists the $F1$-scores and classification accuracies of all shallow learning machines, and the DEPL with the gamma frequency band features. We observed that our method yielded the highest accuracy: 66.23% for valence, 68.50% for arousal (DEAP), and 70.25% for valence and 73.27% for arousal (MAHNOB). At the same time, our method yielded the highest F1-score. We also notice that the ensemble algorithm and the SVM could improve the classification performance against other classical methods. In Fig. 7, the paired $t$-test was adopted to compare whether the performance improvement was significant or not. It can be observed that the DEPL significantly outperformed all the classical shallow learning machines with $p < 0.001$.

TABLE V
PARTICIPANT-AVERAGED CLASSIFICATION PERFORMANCE OF THE DEPL AND SHALLOW LEARNING MACHINES ON TWO DATABASES

| | DEAP | | | | MAHNOB | | | |
|---|---|---|---|---|---|---|---|---|
| | Valence | | Arousal | | Valence | | Arousal | |
| Classifiers | Accuracy | F1-score | Accuracy | F1-score | Accuracy | F1-score | Accuracy | F1-score |
| KNN | 0.5409(3.92e-03) | 0.6152(1.80e-02) | 0.5093(9.18e-03) | 0.5804(2.44e-02) | 0.5431(7.50e-04) | 0.5953(2.50e-03) | 0.5049(7.78e-04) | 0.4977(1.26e-02) |
| NB | 0.5344(8.27e-03) | 0.5937(4.36e-02) | 0.5158(1.48e-02) | 0.5024(6.01e-02) | 0.5227(2.75e-03) | 0.4465(3.71e-03) | 0.5229(5.09e-03) | 0.4318(9.50e-03) |
| RF | 0.5555(7.15e-02) | 0.6940(2.02e-02) | 0.5540(1.04e-01) | 0.6646(3.82e-02) | 0.5506(5.78e-03) | 0.6464(4.78e-02) | 0.5225(7.24e-02) | 0.5833(1.07e-02) |
| AdaBoost | 0.5527(9.15e-02) | 0.7071(1.84e-02) | 0.5467(1.47e-01) | 0.6954(3.23e-02) | 0.5509(7.54e-02) | 0.5998(5.43e-03) | 0.5435(1.39e-01) | 0.5667(1.67e-02) |
| XGBoost | 0.5272(6,12e-02) | 0.6001(4.84e-02) | 0.5129(8.98e-02) | 0.5519(6.21e-02) | 0.5579(5.39e-02) | 0.6022(6.56e-03) | 0.5166(6.53e-02) | 0.5343(1.54e-02) |
| DT | 0.5434(5.84e-03) | 0.5810(1.89e-02) | 0.5234(1.31e-02) | 0.5330(4.57e-02) | 0.5483(1.91e-03) | 0.5974(3.07e-03) | 0.5169(6.43e-03) | 0.5410(1.88e-02) |
| SVM | 0.5531(7.76e-02) | 0.7079(3.84e-02) | 0.5758(9.15e-02) | 0.7174(3.01e-02) | 0.5223(6.88e-02) | 0.5650(1.62e-01) | 0.5338(1.48e-01) | 0.6285(1.88e-01) |
| GBDT | 0.5540(7.09e-02) | 0.7075(8.13e-02) | 0.5519(1.05e-01) | 0.6726(7.57e-02) | 0.5419(4.56e-02) | 0.6899(6.56e-03) | 0.5331(7.31e-02) | 0.6156(1.27e-02) |
| LR | 0.5179(8.59e-03) | 0.5223(5.76e-02) | 0.4878(2.38e-02) | 0.4644(9.11e-02) | 0.5558(3.84e-03) | 0.6116(5.55e-03) | 0.5273(5.15e-03) | 0.5400(2.00e-02) |
| DEPL | **0.6623(5.52e-02)** | **0.7237(4.31e-02)** | **0.6850(5.87e-02)** | **0.7482(3.28e-02)** | **0.7025(5.54e-02)** | **0.7542(2.93e-02)** | **0.7327(6.77e-02)** | **0.7656(1.82e-02)** |

Note: the standard deviation is listed in brackets. The optimal values are shown in boldface.

## D. Comparison with Modern Deep-learning Models

To validate the performance of the DEPL on ER tasks, we introduced several modern deep-learning approaches for comparison, e.g. AlexNet [45] and InceptNet [46]. Moreover, to explore whether the feature fusion from all four frequency bands can improve the ER performance or not, a multi-input model has been designed and denoted as FreqNet. A classical deep-neural network (denoted as ANN) has also been employed for comparison. The network structure of these four deep learning approaches as well as the DEPL is listed in Table VI. It is shown that the DEPL possesses the lowest number of parameters for tuning.

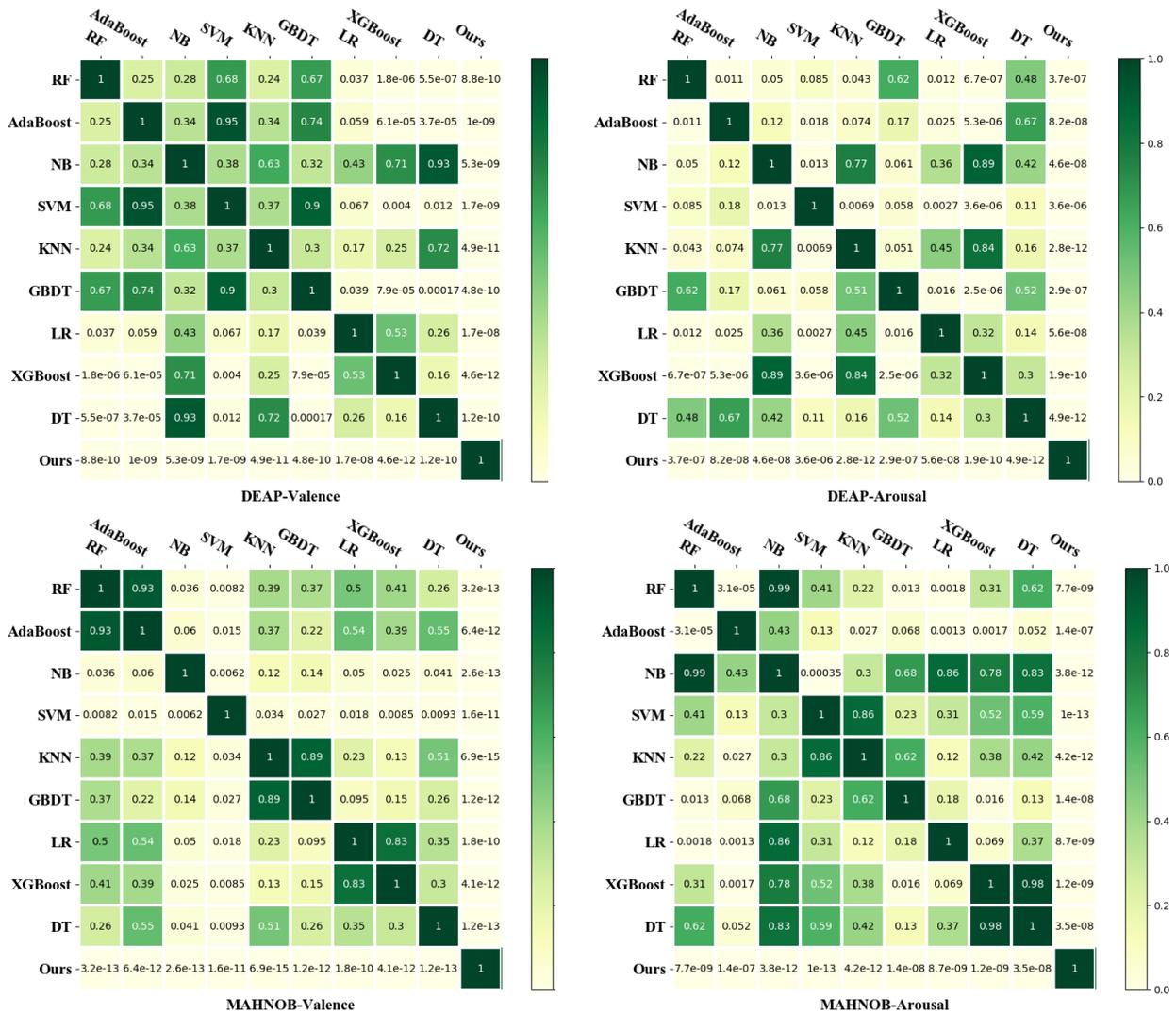

Fig. 7. Heat-map plots of the $p$-values of utilized paired $t$-test between shallow learning machines and the DEPL model.

The average participant accuracies of all five models are listed in Fig. 8. Specifically, the deep ANN penalty strength of the L2 regularization was set to 0.001. The AlexNet, InceptNet, and FreqNet also introduced the SE blocks and the Swish activation function. From the figure, we can observe that the DEPL obtained the highest accuracy. In particular, the classification performance was degraded when more than two convolution-pooling blocks were involved. This indicates that deeper CNN models may not be suitable for EEG feature decoding because of the over-fitting problem. Conversely, the back-propagation-based deep artificial neural network (ANN) achieved the lowest performance. Owing to the fact that the FreqNet model may be too complicated, and the interferences between different frequency bands, the accuracy was also unsatisfactory.

TABLE VI
NETWORK STRUCTURES OF THE MODERN DEEP LEARNING MODELS

| Models | Network structures | Parameters |
|---|---|---|
| ANN | Dense(2000) $\to$ Dropout (0.25) $\to$ Dense(200) $\to$ Dropout (0.5) $\to$ Dense(400) $\to$ Dropout (0.5) $\to$ Softmax(2) | 739,402 |
| AlexNet | Conv(100, 5×5) $\to$ Maxpool(2×2, 2) $\to$ Conv(100, 5×5) $\to$ Maxpool(2×2, 2) $\to$ Conv(50, 3×3) $\to$ Conv(50, 3×3) $\to$ Conv(50, 3×3) $\to$ Dense(120) $\to$ Dense(84) $\to$ Softmax(2) | 363,675 |
| InceptNet | Inception((64,), (96,128), (16,32), (32,)) $\to$ Maxpool(2×2, 2) $\to$ Inception((64,), (96,128), (16,32), (32,)) $\to$ Maxpool(2×2, 2) $\to$ Dense(120) $\to$ Dense(84) $\to$ Softmax(2) | 452,262 |
| FreqNet | Concatenate(Input( $\theta, \alpha, \beta, \gamma$ )) $\to$ Conv(100, 5×5) $\to$ Maxpool(2×2, 2) $\to$ Conv(100, 5×5) $\to$ Maxpool(2×2, 2)) $\to$ Dense(120) $\to$ Dense(84) $\to$ Softmax(2) | 578,902 |
| **DEPL** | Conv(100, 5×5) $\to$ Maxpool(2×2, 2) $\to$ Conv(100, 5×5) $\to$ Maxpool(2×2, 2) $\to$ Dense(120) $\to$ Dense(84) $\to$ Softmax(2) | **152,566** |

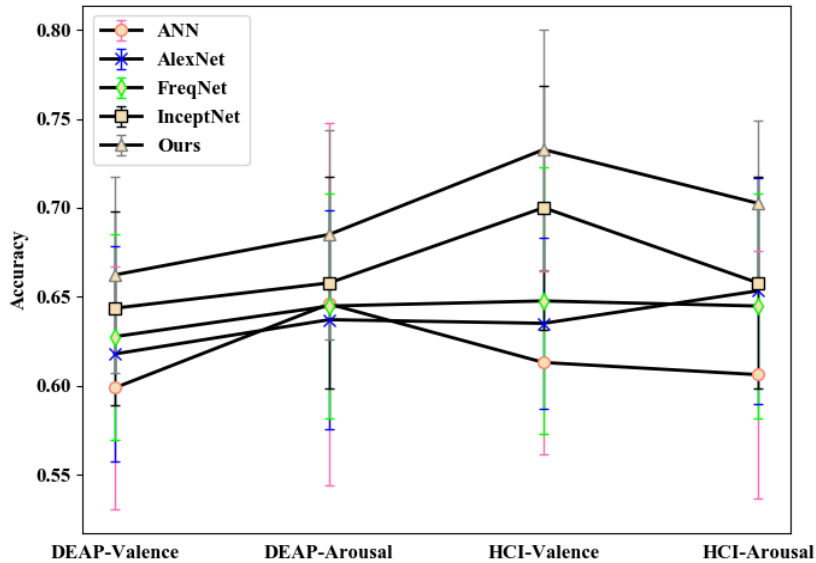

Fig. 8. Error-bar plots of the classification accuracy comparison between deep learning approaches for binary classification into low/high valence and arousal on the DEAP and MAHNOB database, respectively.

*E. Comparison of Computational Complexity*

The computational complexity of the DEPL is compared with several classical learning machines in Table VII with asymptotic notations. It is observed that the running time of the DEPL is drastically higher than those of classical shallow classifiers. The reason behind this is that ensemble learning machines, e.g. RF, adopt divide and conquer principles to design classifier trees. The height of the tree is $\log(n)$ and significantly reduces the increased order of the running time. Conversely, the main time cost of the DEPL is attributed to the procedure of training weights of hierarchical convolution and full connection feature representations. However, compared with several modern DL models, the DEPL has the least number of trainable parameters and achieves the best recognition performance. This part indicates the suitability of the DEPL to tackle the ER tasks with cross-individual training and testing paradigms.

TABLE VII
COMPUTATIONAL COMPLEXITY OF DIFFERENT LEARNING MACHINES

| Training Algorithm | Running time |
|---|---|
| KNN | $O(K*n*d)$ |
| NB | $O(n*d)$ |
| RF | $O(n*\log(n)*d*k)$ |
| AdaBoost | $O(n*\log(n))$ |
| XGBoost | $O(n*\log(n)*d*k)$ |
| DT | $O(n*\log(n)*d)$ |
| SVM | $O(d*n^2)$ |
| GBDT | $O(n*\log(n)*d*k)$ |
| LR | $O(n*d)$ |
| DEPL | $O((n/B)*E*\sum_{l=1}^{D}M^2K^2C_{l-1}C_l)$ |

Note: The term *n* indicates the number of training instances and *d* denotes the dimension of features. For ensemble classifiers, *k* indicates the number of estimators. For the KNN classifier, *K* indicates the number of nearest neighbors. For the DEPL, *M* indicates the size of the feature map output by each convolution kernel, *K* indicates the size of the each convolution kernel, $C_l$ indicates the neuron number of the *l*-th convolutional layer, *E* indicates the number of the epochs, and *B* indicates the batch size.

## VI. DISCUSSION

The competitive performance of the proposed DEPL framework for tackling ER tasks in cross-individual paradigms is mainly attributed to three aspects.

(1) The DEPL is able to eliminate short-term fluctuations and highlight long-term trends in EEG data with the use of dynamical feature smoothing. This allows DEPL to track the temporal dynamics of the affective status over time. (2) The benefits of the DL allow it to be capable to abstract structural data built by locating all EEG channels in a matrix. Deep networks are superior to shallow learning machines because they introduce the layer-wise convolution operations for feature-synthesis, and because they minimize possible inter-individual inconsistencies associated with high-dimensional feature maps. (3) Moreover, the SE-block in conjunction with the Swish activation function efficiently models the interdependencies between the channels, and helps the DEPL converge faster.

21TABLE VIII
CLASSIFICATION PERFORMANCE OF RELATED WORKS OF THE EEG-BASED ER SYSTEMS

| Reference | Database | Classifier | Accuracy (%) for arousal/valence | Feature selection | Type of ER systems |
|---|---|---|---|---|---|
| Atkinson et al. 2016 [24] | DEAP | SVM | 73.06/73.14 | minimum redundancy maximum relevance | individual-dependent |
| Yoon et al. 2013 [25] | DEAP | probabilistic classifier based on Bayes' theorem | 70.10/70.90 | Pearson correlation coefficient | individual-dependent |
| Alhagry et al. 2017 [42] | DEAP | End-to-end deep learning neural networks | 85.65/85.45 | - | individual-dependent |
| Salama et al. 2018 [27] | DEAP | CNN with three-dimensional (3-D) input | 88.49/87.44 | spatiotemporal features mapping | individual-dependent |
| Yang et al. 2018 [28] | DEAP | CNN with 3-D input | 90.24/89.45 | spatiotemporal features mapping | individual-dependent |
| Yang et al. 2018 [48] | DEAP | parallel convolutional recurrent neural network | 91.03/90.80 | spatiotemporal features mapping | individual-dependent |
| Li et al. 2018 [29] | DEAP | SVM | - /59.06 | L1-norm penalty | cross-individual |
| Pandey et al. 2019 [30] | DEAP | deep neural network | 61.25/62.50 | variational mode decomposition | cross-individual |
| Yang et al. 2019 [49] | DEAP | SVM | - /72.00 | significance test and sequential backward selection | cross-individual |
| Yang et al. 2019 [49] | DEAP | SVM | 58.40/57.60 | empirical mode decomposition | cross-individual |
| Rayatdoost et al. 2018 [50] | DEAP | RF | 59.22/55.70 | - | cross-individual |
| **Ours** | **DEAP** | **DEPL** | **66.23/68.50** | **-** | **cross-individual** |
| Rayatdoost et al. 2018 [50] | MAHNOB | RF | 71.25/61.46 | - | cross-individual |
| Soleymani et al. 2012 [22] | MAHNOB | SVM | 52.40/57.00 (3 classes) | - | cross-individual |
| Gao et al. 2015 [51] | MAHNOB | HBN | 63.00/56.90 | Principle component analysis (PCA) | individual-dependent |
| Torres-Valencia et al. 2016 [52] | MAHNOB | SVM | 64.20/63.38 | margin-maximizing feature elimination (MFE) | individual-dependent |
| Momennezhad et al. 2018 [53] | MAHNOB | SVM | 62.10/50.50 | wavelet coefficients | individual-dependent |
| Wiem et al. 2017 [26] | MAHNOB | SVM Gaussian kernel | 63.63/68.75 | feature fusion | individual-dependent |
| Wiem et al. 2017 [26] | MAHNOB | SVM Gaussian kernel | 59.57/57.44 (three classes) | feature fusion | individual-dependent |
| **Ours** | **MAHNOB** | **DEPL** | **70.25/73.27** | **-** | **cross-individual** |

Note: Reported accuracies are elicited by binary classification except those marked by the three classes.

The observation that the gamma frequency band is superior to the performance gained by the theta, alpha, and beta bands, indicates the significance of high-frequency cortical activities in affective computing. The potential reason is attributed to the fact that the gamma waves are heavily involved in the human reasoning and thinking procedures. For the parameter settings of the DEPL, the initial weights learned from different participants are unable to help the DEPL to converge faster owing to the low-migration ability of the current CNN structure. In fact, this implies that the types of the EEG responses are considerable among different individuals involved in the same ER task.

In Table VIII, the performance of the DEPL is compared with recently reported publication findings. In general, the values of the accuracy obtained from the individual-dependent ER systems (exceeded 90%) are far better than those from the cross-individual models (approximately 60%). We found that the DEPL could facilitate much more emotionally relevant feature patterns to achieve better performance, particularly for both the DEAP and MAHNOB databases, compared with the existing methods. However, it should be carefully noted that the size of the training set, data pre-processing steps, and definitions of the emotional classes could be different, even when subjected to the same training and testing paradigms. These factors could significantly affect the final ER accuracy listed in the table.

Through the comparative analysis presented above, the DEPL may contribute to the improved tracking of the temporal dynamics of the affective status from the EEG signals with respect to time to enable cross-individual emotional recognition. The limitations of the current study mainly refer to two points. (1) We only extracted the DE features from raw EEG signals, and did not explore the other entropy measures, such as ApEn [54], fuzzy entropy [55], and multiscale entropy [56]. Different entropy measures may capture informative features from EEG signals in specific emotional states. (2) The DEPL was trained on four frequency bands independently to investigate the performance differences and locate the best bands. However, the proper fusion of the four models may result in a higher generalization capacity. It can also be noted that the cross-individual ER task is still difficult because the accuracy is far from the perfect even for binary affective states.

## VII. Conclusions

In this study, the cross-individual ER system, DEPL, was proposed. The system achieved an outstanding performance with EEG mappings. In the DEPL framework, we exploited dynamic entropy in quantitative EEG measurement over time to extract consecutive differential entropy values to represent the temporal feature profile. A feature-smoothing module has been employed to examine consecutive DE values over time, eliminate short-term fluctuations, and highlight long-term trends among multiple individuals. Specifically, the benefit of the SE-block has been validated with the capability to model the interdependencies across different channel locations. The effectiveness of the DEPL deep-learning framework was also demonstrated by the low number of the tunable parameters compared with modern CNN-based models, and by the acceptable computational complexity.

## Acknowledgment

The authors would like to thank the DEAP and the MAHNOB-HCI teams for providing the multimodal databases to pursue this research project. We also would like to thank Google for providing free-cloud services.

## Author Biographies

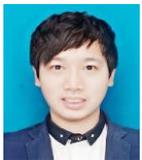

**Xiaolong ZHONG** is with the School of Optical-Electrical and Computer Engineering, University of Shanghai for Science and Technology. His research interests are in the areas of affective computing, coding and decoding theory in brain information processing, analysis of biological neural networks, and computational vision and audition.

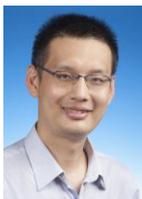

**Zhong YIN** received his PhD in control science and engineering from the East China University of Science and Technology. He was a Lecturer at the School of Optical-Electrical and Computer Engineering, University of Shanghai for Science and Technology, China, from 2015 to 2017. He has served as an Associate Professor at School of Optical-Electrical and Computer Engineering, University of Shanghai for Science and Technology, China, since 2018. His research interests include intelligent human-machine systems, biomedical signal processing and pattern recognition.